\documentclass{article}
\usepackage{spconf,amsmath,graphicx,hyperref}
\usepackage{amssymb}
\usepackage{multirow} 
\usepackage{algorithm}
\usepackage{algorithmic}

\title{A Unified Probabilistic Framework for Dictionary Learning with Parsimonious Activation}
\name{Zihui Zhao$^{1}$, Yuanbo Tang$^{1}$, Jieyu Ren$^{2}$, Xiaoping Zhang$^{1}$,  Yang Li$^{1}$\sthanks{Thanks to National Natural Science Foundation of China (62371270).}}
\address{%
$^{1}$Tsinghua University, Institute of Data and Information,\\
Shenzhen Key Laboratory of Ubiquitous Data Enabling\\[1mm]
$^{2}$University of Chinese Academy of Sciences, Kavli Institute for Theoretical Sciences
}

\usepackage{amssymb}
\begin{document}
\ninept
\maketitle
%

\begin{abstract}
Dictionary learning is traditionally formulated as an $L_1$-regularized signal reconstruction problem. While recent development have incorporated discriminative, hierarchical, or generative structures, most approaches rely on encouraging representation sparsity over individual samples that overlook how atoms are shared across samples, resulting in redundant and sub-optimal dictionaries. We introduce a parsimony promoting regularizer based on the row-wise $L_{\infty}$ norm of the coefficient matrix. This additional penalty encourages entire rows of the coefficient matrix to vanish, thereby reducing the number of dictionary atoms activated across the dataset. We derive the formulation from a probabilistic model with Beta–Bernoulli priors, which provides a Bayesian interpretation linking the regularization parameters to prior distributions. We further establish theoretical 
calculation for optimal hyperparameter selection and connect our formulation to both Minimum Description Length, Bayesian model selection and pathlet learning. Extensive experiments on benchmark datasets demonstrate that our method achieves substantially improved reconstruction quality (with a 20\% reduction in RMSE) and enhanced representation sparsity, utilizing fewer than one-tenth of the available dictionary atoms, while empirically validating our theoretical analysis.
\end{abstract}
\begin{keywords}
Dictionary Learning, Bayesian Modeling, Parsimonious Activation
\end{keywords}
\section{Introduction}

Dictionary learning has long been a fundamental tool in signal processing and machine learning, aiming to find sparse representations of data through an optimization framework~\cite{olshausen1997sparse}. Traditional methods focus on minimizing reconstruction error with sparsity-inducing regularization, typically expressed as:  
\begin{equation}
\min_{D,R} \left[ \|X - D R\|_F^2 + \lambda \|R\|_1 \right]
\end{equation}
where $ X \in \mathbb{R}^{d \times m}$ is the data matrix, $D \in (0,1)^{d \times n}$ is the dictionary matrix, and $R \in \mathbb{R}^{n \times m}$ is the coefficient matrix. Here, $d$ denotes the feature dimension, $m$ is  the number of samples, and $n$ is the number of dictionary atoms. The indices $i$ (from 1 to $n$) and $j$ (from 1 to $m$) refer to specific samples and dictionary atoms, respectively.

Recent advances extend the basic objective in different directions~\cite{aharon2006k}. Supervised/discriminative dictionary-learning methods jointly learn dictionaries and classifiers so that sparse codes are also predictive (e.g. label-consistent K-SVD and related methods~\cite{jiang2013label, zhou2016bilevel, guo2016discriminative, shao2020label, wang2017cross}). Deep or hierarchical dictionary-learning approaches stack multiple dictionary-learning layers to learn multi-level representations. Methods that combine latent-variable models, variational autoencoders (VAEs) or nonnegative decompositions have also been proposed for tasks such as speech enhancement and structured representation learning~\cite{leglaive2019semi}. 

More recently, deep dictionary learning methods have emerged which optimize dictionaries and sparse codes to learn atoms that support not only faithful signal reconstruction but also improved performance on downstream tasks. Such approaches have further been demonstrated to enhance interpretability and to yield semantically coherent or biologically meaningful representations across diverse application domains~\cite{makelov2024towards,kang2024deep}.


Previous dictionary learning approaches typically impose an $L_1$ penalty on overall sparsity, thereby treating all dictionary atoms uniformly. In contrast, our work introduces an additional constraint designed to achieve accurate reconstruction while minimizing the number of activated atoms. Drawing inspiration from the Beta–Bernoulli prior, we employ this sparse distribution to selectively activate only a small subset of the dictionary. Unlike prior Bayesian formulations that rely on complicated variational estimation to learn the intractable   posterior~\cite{shah2015empirical,paisley2009nonparametric}, our approach leads to a simple formulation that can be solved using projected gradient descent.

Specifically, we propose a probabilistic dictionary learning framework which leads to   a  three-part optimization objective:
\begin{equation}
\label{eq:objective}
\min_{D,R} \left[ \|X - D R\|_F^2 + \lambda_1 \|R\|_1 + \lambda_2 \sum_{i=1}^m \|r_i\|_\infty \right]
\end{equation}
We derive this objective from maximum a posterior (MAP) estimation on a probabilistic  model with a Beta–Bernoulli prior on both activations and coefficient magnitudes, providing a rigorous statistical justification for the inclusion of $L_1$ and row-wise $L_{\infty}$ penalties. Specifically, our probabilistic model applies the infinity norm ($\sum_{i=1}^n \|r_i\|_\infty$) to encourage rows in $R$ to be (nearly) zero, thus reducing the number of atoms actually used across the dataset. Furthermore, we offer theoretical guarantees for selecting the optimal $\lambda_1$ and $\lambda_2$, based on explicit bounds on reconstruction error and sparsity levels, thereby mitigating the common dependence on heuristic hyperparameter tuning. 

In addition, our framework unifies different perspectives on dictionary learning algorithm design. First, we show that the proposed objective can be interpreted from a Minimum Description Length (MDL) perspective~\cite{rissanen1978modeling} and is asymptotically equivalent to Bayesian model selection~\cite{schwarz1978estimating}. We further establish connections to the pathlet learning problem~\cite{buchin2011detecting, alix2023pathletrl, chen2013pathlet,li2016knowledge,tang2023explainable}, which seeks to identify the minimal set of representative pathlets that can reconstruct original trajectories using as few components as possible. To the best of our knowledge, our work provides the first rigorous theoretical justification for this optimization approach. 

Finally, rigorous experiments on standard image reconstruction benchmarks validate both the reconstruction performance of our method and the effectiveness of our theory-driven  hyperparameter estimation method. 

\section{The Bayesian Framework}
\subsection{Probabilistic Model}

We develop a Bayesian framework that naturally leads to our three-part objective. Assume coefficients follow independent Beta distributions:
\begin{equation}
R_{ij} \sim \text{Beta}(1, \beta), \quad \beta > 1
\end{equation}
with probability density function:
\begin{equation}
p(R_{ij}) = \beta(1 - R_{ij})^{\beta-1}, \quad R_{ij} \in [0,1]
\end{equation}
Define latent activation variables $z_i$ in $\{0,1\}$with:
\begin{equation}
z_i \mid R_i \sim \operatorname{Bernoulli}\left( \frac{1}{1 + \exp\left( \gamma (\max_j R_{ij} - \delta) \right)} \right)
\end{equation}
where $\delta \in [0,1]$ denotes the activation threshold and $\gamma > 0$ controls the sharpness of the decision. 
The term $\max_j R_{ij}$ is employed to capture the maximum value in the $j$-th row of $R$, which serves as an indicator of whether a specific dictionary atom is activated when combined with the threshold $\delta$.
The data generation process is:
\begin{equation}
X_i = D R_i + \epsilon_i, \quad \epsilon_i \sim \mathcal{N}(0, \sigma^2 I)
\end{equation}
where $D$ is a learnable model parameter, rather than being modeled as a random variable as in previous approaches.

\subsection{Posterior Inference}
With the above-mentioned priors, we have the joint probability distribution:
\begin{equation}
P(X, R, z ; D) = \prod_{i=1}^N \left[ P(X_i | R_i; D) \cdot \prod_{j=1}^K p(R_{ij}) \cdot P(z_i | R_i) \right]
\end{equation}
Taking the negative log-posterior, we have:
\begin{equation}
\label{eq:logP}
\begin{aligned}
-\log P(R,z | X;D) = & \frac{1}{2\sigma^2} \sum_{i=1}^N \|X_i - D R_i\|^2 \\
& - \sum_{i=1}^m \sum_{j=1}^n \log p(R_{ij})  - \sum_{i=1}^m \log P(z_i | R_i) + C
\end{aligned}
\end{equation}

\subsection{Regularization Equivalence}
In this section we will show how our assumed priors can derive the optimization objective~\ref{eq:objective} under the MAP. We will show that the second term and third term are equivalent with the $L_1$ regularization and the $L_{\infty}$ regularization. 

For small $R_{ij}$ values ($\beta > 1$), we have:
\begin{equation}
-\sum_{i,j} \log p(R_{ij}) \approx (\beta-1) \|R\|_1
\end{equation}
which is based on $-\log p(R_{ij}) \approx (\beta-1)R_{ij}$, this is the $L_1$ regularization in Eq.~\ref{eq:objective}. 

For the infinity norm regularization term, we first employ a constructor function:
\begin{equation}
\phi_\gamma(u) = \log(1 + e^{\gamma u})
\end{equation}
Then apply it on the third term in Eq.~\ref{eq:logP}: 
\begin{equation}
-\log P(z_i | R_i) = \phi_\gamma(\max_j R_{ij} - \delta)
\end{equation}
As $\gamma \rightarrow \infty$
\begin{equation}
\lim_{\gamma \to \infty} \phi_\gamma(u) = \begin{cases} 
0 & u < 0 \\
\infty & u > 0 
\end{cases}
\end{equation}
This imposes the constraint $max_j R_{ij} \le \delta$ for all $i$. The complete optimization objective becomes:
\begin{equation}
\min_{D,R} \left[ \frac{1}{2\sigma^2} \|X - D R\|_F^2 + \lambda_1 \|R\|_1 + \lambda_2 \sum_i \phi_\gamma(\max_j R_{ij} - \delta) \right]
\end{equation}
With $\sigma$ = 1 (achieved by normalizing $X$) and $\gamma \rightarrow \infty$ , we obtain our novel objective:
\begin{equation}
\label{eq:novel-obj}
\min_{D,R} \left[ \|X - D R\|_F^2 + \lambda_1 \|R\|_1 + \lambda_2 \sum_{i=1}^m \max_j |r_{ij}| \right]
\end{equation}
where $\lambda_1$ = $\beta-1$ and $\lambda_2$ = 1.

\subsection{Theoretical Analysis}
Assume there exists a dictionary subset $S$ (important atoms) such that:
\begin{equation}
\|X - D_S R_S\|_F^2 \leq \epsilon
\end{equation}
For atom $d_k$ not in S, we want their coefficients $r_k = 0$. The condition for setting these to zero is:
\begin{equation}
\begin{aligned}
\Delta f &= f(R_{-k}) - f(R) \\
&= 2 \sum_i e_i^\top d_k r_{ik} + \|d_k r_k^\top\|_F^2 - \lambda_1 \|r_k\|_1 - \lambda_2 \max_i |r_{ik}| < 0
\end{aligned}
\end{equation}
where $\epsilon = X - D_S R_S$ is the reconstruction error.
If $|e_i^\top d_k| \le \eta$ for all $i$, and dictionary atoms are normalized ($||d_k||_2 = 1$), then a sufficient condition for $\Delta f < 0$ is:
\begin{equation}
\lambda_1 \|r_k\|_1 + \lambda_2 \max_i |r_{ik}| \geq (2\eta + \delta) \|r_k\|_1
\end{equation}
where $\delta$ is an upper bound on $|r_{ij}|$.

As established in Eq.~\ref{eq:novel-obj}, $\lambda_2 = 1$ is the natural choice. Assume that $ \beta > 1$ which controls coefficient sparsity (prior parameter), $ \eta$ is the upper bound of reconstruction error-atom correlation and $ \delta$ is the coefficient upper bound (activation threshold). For $\lambda_1$, we have:
\begin{equation}
\lambda_1 = \max(\beta - 1, 2\eta + \delta)
\end{equation}

The correlation bound $\eta$ can be estimated as:
\begin{equation}
\eta = \max_{i,k} |e_i^\top d_k| \approx \frac{\|e\|_F}{\sqrt{m}} \quad \text{(assuming error-atom independence)}
\end{equation}

\subsection{Dictionary Learning Algorithm}


Our dictionary learning framework employs alternating optimization with projected gradient descent to jointly learn the dictionary, the coefficients and the associated regularization as shown in Algorithm.~\ref{alg:dictionary_learning}. In practice, we first estimate the Beta distribution parameters of $R$ on the training dataset, which subsequently inform the selection of the optimal coefficient settings.

\begin{algorithm}
\caption{Dictionary Learning with Process}
\label{alg:dictionary_learning}
\begin{algorithmic}[1]
\REQUIRE Data matrix $\mathbf{X}$, maximum iterations $T$
\ENSURE Learned dictionary $\mathbf{D} \in \mathbb{R}^{d \times n}$

\STATE $\mathbf{X}_{\text{norm}} \gets (\mathbf{X} - \boldsymbol{\mu}) \oslash \boldsymbol{\sigma}$
\STATE // Initialize dictionary and representation coefficients
\STATE $\mathbf{D} \gets \text{random matrix with values in } [0, 1]$
\STATE $\mathbf{R} \gets \text{initialize with Beta distribution}$
\FOR{$t = 1$ \TO $T$}
    \FOR{$i = 1$ \TO $k_r$}
        \STATE $\mathbf{R} \gets \arg\min_{\mathbf{R}} \|\mathbf{X}_{\text{norm}} - \mathbf{D}\mathbf{R}\|_F^2 + \lambda_1\|\mathbf{R}\|_1 + \lambda_\infty\sum_j\max_i|\mathbf{R}_{ij}|$
        
    \ENDFOR
    \FOR{$j = 1$ \TO $k_d$}
        \STATE $\mathbf{D} \gets \arg\min_{\mathbf{D}} \|\mathbf{X}_{\text{norm}} - \mathbf{D}\mathbf{R}\|_F^2$
    \ENDFOR  
\ENDFOR

\STATE \RETURN $\mathbf{D}$
\end{algorithmic}
\end{algorithm}

\subsection{MDL Interpretation: A Theoretical Perspective}

The MDL principle provides an information-theoretic perspective on our framework. The total description length consists of:
\begin{equation}
\text{TDL} = L(H) + L(D|H)
\end{equation}
where $H$ denotes the hypothesis, comprising both the model structure and parameters, while $L(H)$ represents the model description length and $L(X|H)$ the data description length. In the Bayesian learning context, the MDL principle seeks to identify the hypothesis that minimizes the total description length (both data and model description length). 
In our dictionary learning context:
\begin{itemize}
    \item $L(X|H) =||R||_1 + \epsilon$  (data description length) 
    \item $L(H) \propto \Sigma_{i=1}^m ||r_i||_{\infty}$ (model description length)
\end{itemize}
where $\epsilon$ is the reconstruction error. From the Bayesian perspective $P(H|X) = \frac{P(X|H)P(H)}{P(X)}$, taking negative logarithms: 
\begin{equation}
-\log P(H|X) = -\log P(X|H) - \log P(H) + \log P(X)
\end{equation}

The $-log P(\cdot)$ corresponds (up to constants) to optimal code lengths (or description lengths). Thus we can identify:
\begin{equation}
L(H \mid X) := -\log P(X \mid H) - \log P(H)
\end{equation}
Hence, our estimator with $L_\infty$ norm is equivalent to minimizing the total description length:
\begin{equation}
    \hat{H}_{MAP} = \arg\max_H P(H \mid X) = \arg\min_H \big( L(X \mid H) + L(H) \big)
\end{equation}
This equation shows that our proposed $L_\infty$ regularization performs as the model description length term and the $L_1$ term with reconstruction error correspond to the data description length in the MDL.

\subsection{Connection to Pathlet Learning}

Our proposed framework also establishes a formal connection to pathlet learning through infinity norm regularization. In pathlet learning, trajectories are represented as few concatenation of path segments (pathlets), based on the intuition that important pathlets are frequently shared among most trajectories~\cite{chen2013pathlet,li2016knowledge,tang2023explainable}. While this approach was initially motivated by intuition, our framework offers a mathematically explainable formulation. Most importantly, by incorporating a Bayesian interpretation through the specified prior distribution \(P(R_{ij})\) and activation function \(P(z_i | R_i)\), our method establishes a probabilistic foundation for pathlet representation. This offers a statistical justification that was previously absent in the literature. Furthermore, we provide a general continuous-domain formulation, whereas prior work in pathlet learning can be viewed as a discrete instantiation of our framework, where both the dictionary and the encoded representations are discrete.


\section{Experiments}

\subsection{Experimental Framework}

Our experimental evaluation focuses on validating the theoretical contributions of our framework, particularly the Bayesian interpretation and the parameter selection guidelines. We conducted experiments on CIFAR-100~\cite{krizhevsky2009learning} and SVHN~\cite{netzer2011reading} to demonstrate the practical implications of our theoretical framework. Each dataset containing 60,000 color images of size 32 $\times$ 32. All images were converted to grayscale, and non-overlapping 8 $\times$ 8 patches were extracted for training and evaluation. The dictionary size was fixed at 128 atoms across all experiments. During training, 20,000 patches were randomly selected, while an additional 2,000 distinct patches were reserved for testing and performance evaluation.

\subsection{Comparative Reconstruction Performance}

While our primary focus is theoretical validation, we include a concise comparison of reconstruction performance to demonstrate practical utility. Table~\ref{tab:reconstruction_comparison} shows that our method achieves competitive reconstruction quality while providing theoretical guarantees.

\begin{table}[htbp]
\centering
\caption{Reconstruction performance comparison on CIFAR-100 and SVHN. Our method achieves competitive results while providing theoretical guarantees for parameter selection.}
\label{tab:reconstruction_comparison}
\begin{tabular}{lcccc}
\hline
\textbf{Dataset} & \textbf{Method} & \textbf{RMSE} & \textbf{PSNR} & \textbf{SSIM} \\
\hline
\multirow{3}{*}{CIFAR-100}
& Standard KSVD~\cite{aharon2006k}      & 0.152 & 17.87 & 0.472 \\
& DDL~\cite{tariyal2016deep}  & 0.132 & 18.07 & 0.514 \\
& Greedy-DDL~\cite{tariyal2016greedy}  & 0.123 & 18.32 & 0.519 \\
& $L_1$-Regularized DL  & 0.112 & 18.98 & 0.545 \\
& Proposed Method    & 0.099 & 20.56 & 0.568 \\
\hline
\multirow{3}{*}{SVHN}
& Standard KSVD      & 0.147 & 16.34 & 0.449 \\
& DDL  & 0.124 & 18.01 & 0.527 \\
& Greedy-DDL  & 0.118 & 19.33 & 0.539 \\
& $L_1$-Regularized DL  & 0.094 & 20.46 & 0.559 \\
& Proposed Method    & 0.090 & 20.88 & 0.583 \\
\hline
\end{tabular}
\end{table}

\subsection{Ablation Studies}


We conducted comprehensive ablation studies to evaluate the contribution of each component in our framework. Table~\ref{tab:ablation_study} shows the performance of different configurations:

\begin{table}[htbp]
\centering
\caption{Ablation study demonstrating the contribution of each component in our framework.}
\label{tab:ablation_study}
\begin{tabular}{lccc}
\hline
\textbf{Configuration} & \textbf{RMSE} & \textbf{PSNR} & \textbf{SSIM}  \\
\hline
Full framework & 0.099 & 20.56 & 0.568   \\
w/o Infinity norm  & 0.112 & 18.98 & 0.545   \\
w/o $L_1$  & 0.108 & 19.17 & 0.549   \\
w/o $L_1$ \& Infinity norm  & 0.117 & 18.49 & 0.525   \\
\hline
\end{tabular}
\end{table}

The results show that both regularization terms contribute significantly to the performance, with the infinity norm term providing particularly important theoretical properties.



\subsection{Theoretical Parameter Validation}

According to our theoretical framework, the optimal parameters should satisfy $\lambda_1 = \beta - 1$ and $\lambda_2 = 1$, where $\beta$ is the from the Beta prior distribution of $R$ matrix. We compared the theoretically recommended parameters with empirically optimal values found through extensive grid search. Table~\ref{tab:parameter_validation} shows the close alignment between theoretical predictions and empirical optima.

\begin{table}[htbp]
\centering
\caption{Validation of theoretical parameter selection guidelines. Our theoretical framework accurately predicts optimal parameter values.}
\label{tab:parameter_validation}
\begin{tabular}{lccc}
\hline
\textbf{Parameter} & \textbf{Theoretical} & \textbf{Empirical} & \textbf{Relative Error $\%$}\\
\hline
$\lambda_1$ & 1.31 & 1.33 &2.3\\
$\lambda_2$ & 1.00 &   1.01 & 1.0\\
\hline
\end{tabular}
\end{table}

\begin{figure}
    \centering
    \includegraphics[width=0.9\linewidth]{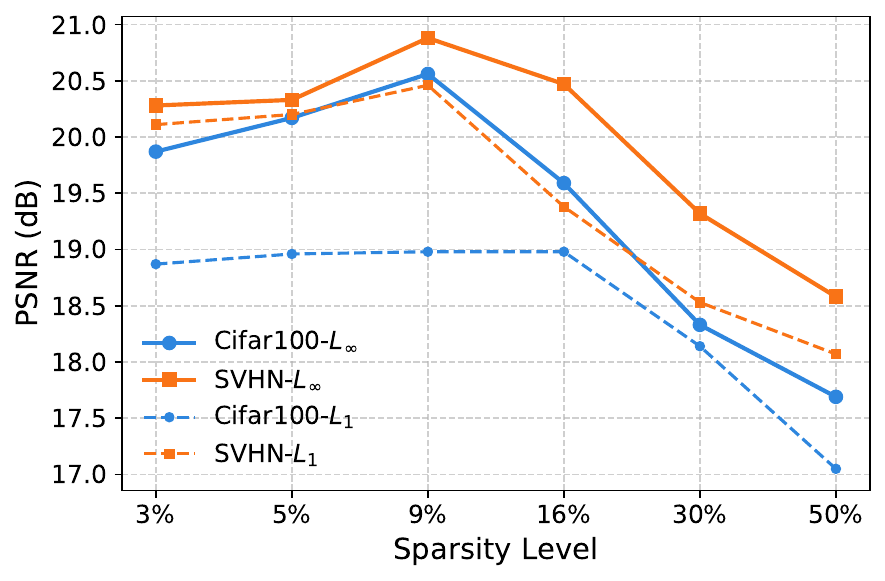}
    \caption{PSNR comparison on Cifar100 and SVHN datasets with varied sparsity level.}
    \label{fig:sparsity_sensitivity}
\end{figure}

\subsection{Minimum Description Length Validation}


Figure~\ref{fig:sparsity_sensitivity} shows the sensitivity of reconstruction error to sparsity variations. Compared with the $L_1$-regularized approach, our method achieves superior reconstruction performance under a sparser dictionary learning regime, while requiring fewer dictionary atoms. 

\begin{figure}
    \centering
    \includegraphics[width=0.9\linewidth]{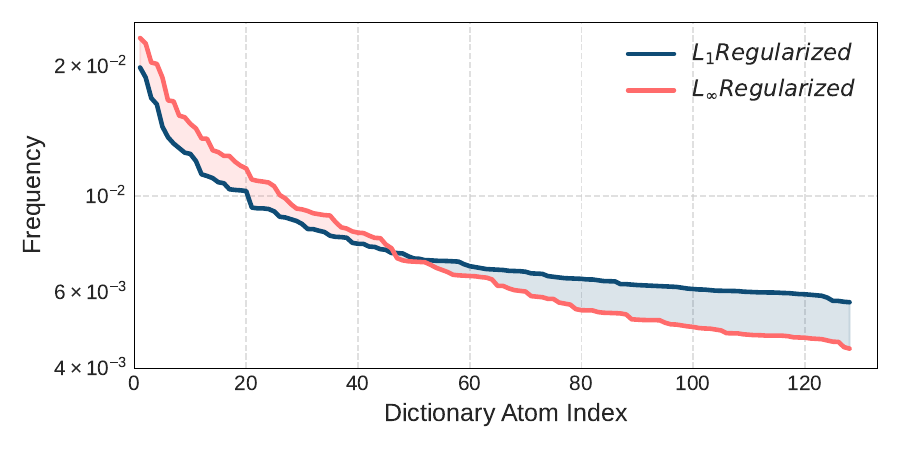}
    \caption{We compare the frequency of atom utilization on Cifar100 between the $L_\infty$-regularized and $L_1$-regularized dictionaries. }
    \label{fig:atoms_sorted}
\end{figure}

Figure~\ref{fig:atoms_sorted} presents a comparison of dictionary atom usage frequencies between the $L_\infty$- and $L_1$-regularized dictionaries, evaluated under identical parameter configurations and averaged over the CIFAR‑100 test set. The usage in our model is concentrated on a smaller subset of the original dictionary, suggesting that the proposed regularization on $R$ encourages a more compact and efficient representation. This also corresponds to a reduced description length, as the reconstruction process involves fewer dictionary atoms overall. Consequently, our method achieves a better trade-off between model complexity and reconstruction error, effectively minimizing the total description length as predicted by our theoretical analysis.

\section{Conclusion}
This paper has presented a unified theoretical framework for regularized dictionary learning that incorporates parsimonious activation constraints through infinity norm regularization. By establishing connections between Bayesian inference, optimization theory and minimum description length principles, we have provided a rigorous mathematical foundation for controlling sample-wise sparsity patterns in dictionary representations. Our key innovation lies in introducing infinity-norm regularization to address a fundamental limitation of traditional dictionary learning methods, which lack explicit mechanisms to constrain the maximum number of atoms activated per sample.

The primary limitation of our current work lies in its relatively straightforward architecture. Nevertheless, the focus of this paper is on formulating an explainable Bayesian framework that learns a compact dictionary while providing a principled approach for optimal parameter selection. Extending this framework with deep neural networks and latent representations to enhance expressive power is an important direction for future research. While this paper only include evaluations on image datasets, we will also explore broader applications, such as constructing multi-modal dictionaries with shared atom activation pattern across modalities.
\vfill\pagebreak

\bibliographystyle{IEEEbib}
\bibliography{refs}

\end{document}